\documentclass[11pt]{article}

\usepackage[final]{acl}

\usepackage{times}
\usepackage{latexsym}

\usepackage[T1]{fontenc}

\usepackage[utf8]{inputenc}

\usepackage{microtype}

\usepackage{inconsolata}

\usepackage{graphicx}

\usepackage{amsmath}
\usepackage{amssymb}
\usepackage{amsfonts}
\usepackage{bm} 
\usepackage{booktabs}    
\usepackage{multirow}    
\usepackage{array}       
\usepackage{enumitem}

\title{From Attenuation to Attention: Variational Information Flow Manipulation for Fine-Grained Visual Perception}

\author{
    Jilong Zhu\textsuperscript{\rm 1,2,3},
    Yang Feng\textsuperscript{\rm 1,2,3}\thanks{Corresponding author.} \\
    \textsuperscript{\rm 1} Key Laboratory of Intelligent Information Processing, \\
    Institute of Computing Technology, Chinese Academy of Sciences (ICT/CAS) \\
    \textsuperscript{\rm 2} State Key Laboratory of AI Safety, Institute of Computing Technology, \\
    Chinese Academy of Sciences (ICT/CAS) \\
    \textsuperscript{\rm 3} University of Chinese Academy of Sciences, Beijing, China \\
    \texttt{\href{mailto:zhujilong22s@ict.ac.cn}{zhujilong22s@ict.ac.cn},  \href{mailto:fengyang@ict.ac.cn}{fengyang@ict.ac.cn}}
}

\begin{document}
\maketitle
\begin{abstract}
While Multimodal Large Language Models (MLLMs) have demonstrated impressive capabilities in general visual understanding, they frequently falter in fine-grained perception tasks that require identifying tiny objects or discerning subtle visual relationships. We attribute this limitation to Visual Attenuation: a phenomenon where sparse fine-grained visual signals are prematurely suppressed or diluted by dominant textual tokens during network propagation, resulting in a ``loss of focus'' during the deep-level decision-making process. Existing input-centric solutions fail to fundamentally reverse this intrinsic mechanism of information loss. To address this challenge, we propose the Variational Information Flow (VIF) framework. Adopting a probabilistic perspective, VIF leverages a Conditional Variational Autoencoder (CVAE) to model the visual saliency relevant to the question-answer pair as a latent distribution. As a plug-and-play module, VIF can be integrated into existing architectures. Extensive evaluations across diverse benchmarks—covering General VQA, fine-grained perception, and visual grounding—demonstrate that VIF yields competitive improvements over previous methods, validating its effectiveness in enhancing the fine-grained perception of MLLMs. Codes are
available at \url{https://github.com/ictnlp/VIF}.
\end{abstract}


\section{Introduction}

In recent years, Multi-modal Large Language Models (MLLMs), represented by the LLaVA~\cite{liu2023visual,huang2025hires}, InternVL~\cite{chen2024internvl, wang2025internvl3_5}, and QwenVL~\cite{Qwen-VL, Qwen2-VL, Qwen2.5-VL, Qwen3-VL} series, have demonstrated remarkable proficiency in general visual understanding and reasoning. However, as the focus shifts from macroscopic scene description to \textbf{Fine-Grained Visual Perception}, a critical bottleneck becomes apparent. Current MLLMs often underperform when tasked with discerning minute objects or interpreting subtle visual relationships, limiting their broader applicability. 

\begin{figure}[t]
    \centering
    \includegraphics[width=\linewidth]{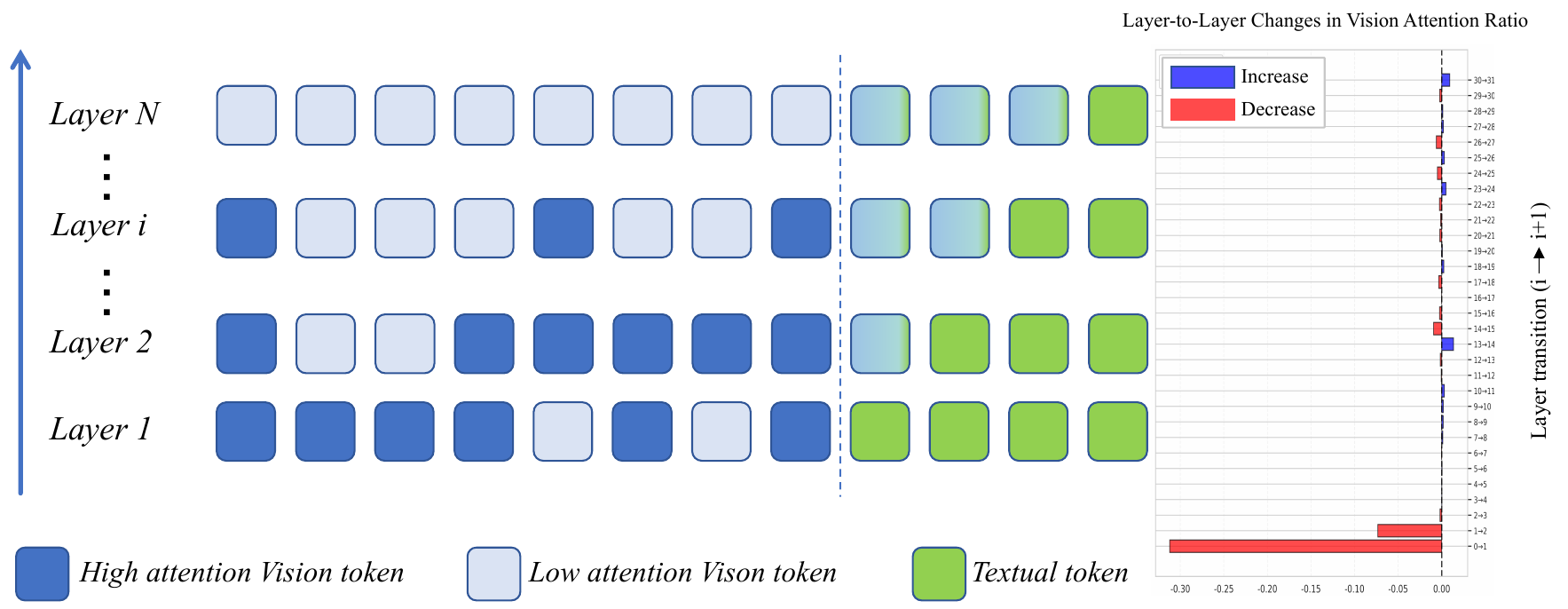} 
    \caption{\textbf{Illustration of Visual Attenuation.} \textbf{(Left)} Schematic view showing visual tokens (blue) fading as they propagate through deep layers, while textual tokens (green) dominate. \textbf{(Right)} Quantitative layer-to-layer changes in vision attention ratio, averaged over 500 randomly sampled instances. The sharp drop in early layers indicates a premature loss of visual details.}
    \vspace{-2ex}
    \label{fig:visual_attenuation}
\end{figure}

This performance degradation stems from the information flow mechanism of MLLMs according to our analysis experiments (details will be shown in Section~\ref{sec:analysis}). In the current information flow mechanism of MLLMs, visual and textual inputs are fed into the model independently. After undergoing initial processing in the shallow layers, semantic representations for both modalities are learned at the middle layers. However, as information ascends to the deep layers, textual representations continuously absorb visual information, leading to a gradual attenuation of visual signals. 
This structural bias ultimately causes the model to prioritize textual representations over visual ones during the downstream textual response generation. Notably, textual representations typically integrate only those visual cues that are relevant to textual instructions, whereas the visual information critical for response generation is often not explicitly reflected in such instructions. Further experimental investigations reveal that, during the textual response generation phase, visual representations not only receive a marginal proportion of attention weights but also tend to exhibit a uniform distribution, ultimately resulting in the defocusing of visual information.

Building on the aforementioned findings, enabling fine-grained image understanding requires assigning greater attention to response-relevant visual regions during the generation process. Furthermore, attention over visual representations should be selectively prioritized rather than uniformly distributed. However, these objectives are challenged given the inherent architectural constraints of existing MLLMs. First, MLLMs lack access to response information during inference, making it difficult to identify visual information relevant to the target response. Second, the current formulation of attention distribution lacks explicit constraints, which constitutes a critical bottleneck in guiding the model to focus on specialized visual cues.

Existing research(e.g., Dense Connector~\cite{yao2024dense}, MMFuser~\cite{cao2024mmfuser}) on fine-grained visual perception primarily focuses on enhancing the fidelity of visual inputs. While these methods effectively augment the amount of visual information at the input stage, they still suffer from visual attenuation during cross-modal transmission where visual representations receive insufficient attention, and textual representations fail to capture the fine-grained visual details essential for generating accurate responses.

On these grounds, we introduce a novel \textbf{Variational Information Flow (VIF)} framework, designed to regulate the flow of visual information for seamless adaptation to both fine-grained and macroscopic visual tasks. To address the challenge of identifying response-correlated visual information that cannot be directly inferred from textual instructions, we employ variational inference to guide the model toward essential visual cues. 
In the training phase, we supply the ground-truth response, which facilitates the derivation of two heterogeneous attention distributions over visual representations: a posterior attention distribution, which takes both textual instructions and responses as query, and a prior attention distribution conditioned solely on textual instructions. By aligning these two distributions, VIF equips the MLLM with the capability to prioritize response-relevant visual attention during inference, even when only textual instructions are provided as queries. 

To further enhance the model’s focus on critical visual cues, we formalize the attention distribution using a Gaussian Mixture Model (GMM), where each Gaussian component in this model serves as a spotlight, re-activating the key visual information that might otherwise be neglected in the response generation process. Extensive experiments demonstrate that VIF, as a plug-and-play module, simultaneously supports fine-grained perception and broader macroscopic tasks, consistently outperforming existing input feature augmentation methods across both general and fine-grained benchmarks, validating the efficacy of our probabilistic information reconstruction paradigm.

Our main contributions are as follows:

i. We identify {Visual Attenuation} in the deep layers of MLLMs, revealing both a quantitative decline in attention weights and a structural collapse in spatial focus, highlighting the limitations of methods that only enhance input features.

ii. We propose the VIF framework, which employs CVAE-based posterior learning to dynamically reconstruct visual attention and capture task-specific visual saliency through robust probabilistic modeling.

\section{Preliminary Experiments}
\label{sec:analysis}

To further investigate the Visual Attenuation, we conducted an in-depth statistical analysis of the internal attention dynamics of MLLMs based on 500 randomly sampled instances. 

\subsection{Vision Attention Ratio Analysis}
\begin{figure*}[t]
    \centering
    \includegraphics[width=0.95\linewidth]{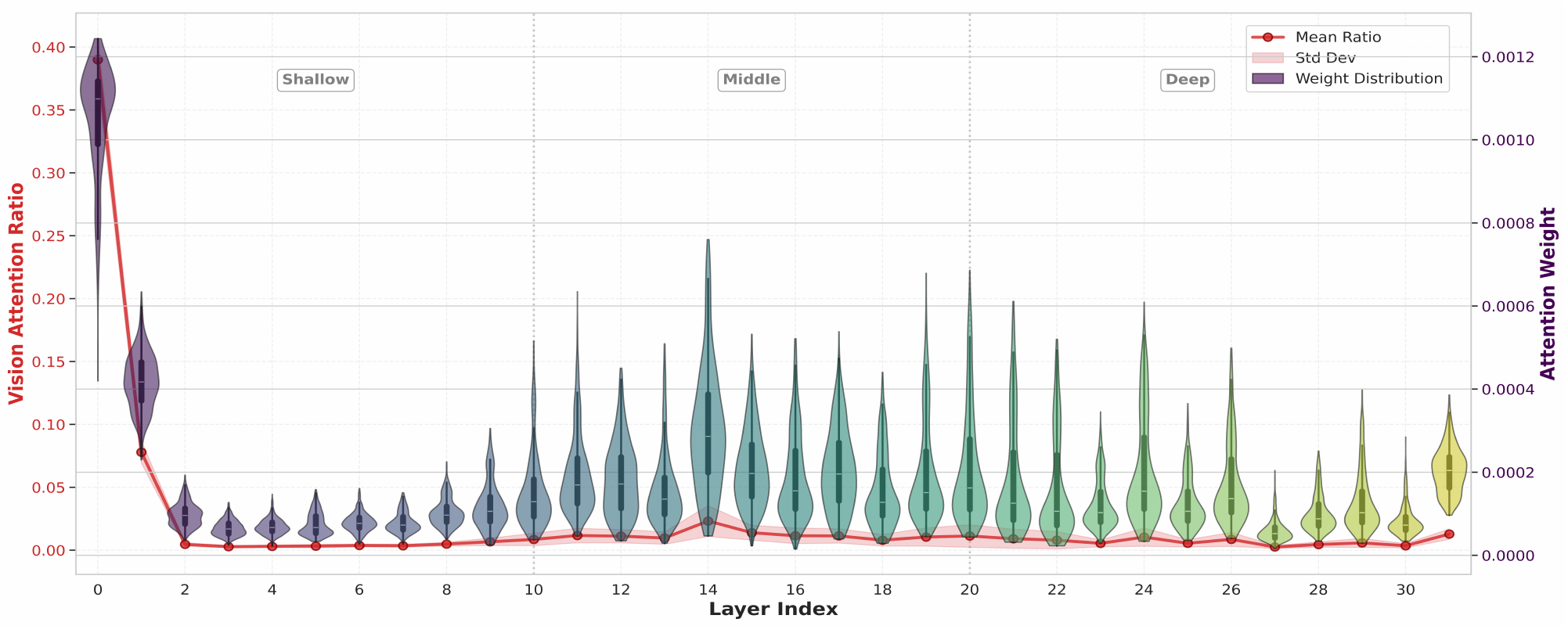}
    \vspace{-2ex}
    \caption{\textbf{Layer-wise visual attention distribution.} The red curve denotes the mean vision attention ratio across layers, which decreases sharply with depth. The violin plots illustrate the distribution of visual attention weights at each layer, with the long-tailed distribution shrinking in the deep layers.}
    \label{fig:attention_distribution}
\end{figure*}

As shown in the right panel of Figure~\ref{fig:visual_attenuation} and the red curve in Figure~\ref{fig:attention_distribution}, we observe a pervasive phenomenon of Visual Attenuation: although visual tokens initially command high attention weights in shallow layers, their influence suffers a precipitous decline in deep layers. Critical visual cues are overwhelmed by dominant textual tokens and prematurely discarded, hindering the model's ability to discern subtle details.

\begin{figure*}[t]
    \centering
    \includegraphics[width=0.99\linewidth]{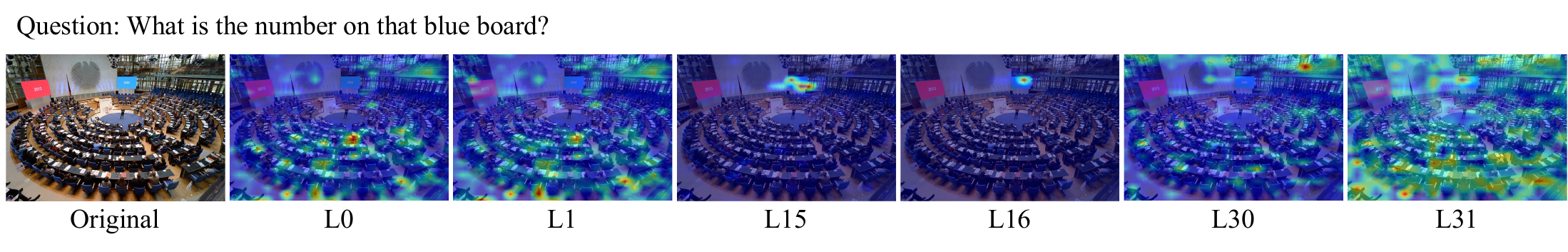}
    \vspace{-2ex}
    \caption{\textbf{Visualization of layer-wise visual attention.} \textbf{(Left)} Shallow layers (L0-L1) capture dense contextual information with broad coverage. \textbf{(Middle)} Middle layers (L15-L16) successfully converge on key semantic regions (the board), exhibiting sparse and focused attention. \textbf{(Right)} In deep layers (L30-L31), this focus deteriorates into a diffuse and disordered state.}
    \vspace{-2ex}
    \label{fig:attention_visualization}
\end{figure*}

\subsection{Vision Attention Distribution Analysis}
Further spatial analysis reveals that, beyond the mean attention ratio decline, the internal allocation structure of visual attention also collapses. As illustrated by the violin plots in Figure~\ref{fig:attention_distribution}, the long-tailed distribution shrinks in deep layers, with attention weights converging to homogenized low values that approximate a uniform spatial distribution. This indicates a reduced capacity for selectively activating key visual tokens.

This observation is further spatially corroborated by the visualizations in Figure~\ref{fig:attention_visualization}: while the model successfully converges on key semantic regions (e.g., the screen) in the middle layers (L15-L16), this focused state rapidly deteriorates into a diffuse and disordered distribution in the deep layers (L30-L31). 
Thus, beyond the mere neglect of visual signals in deep layers, Visual Attenuation also manifests as a spatially structural collapse.

This dual degradation in preference magnitude and spatial structure indicates that MLLMs not only \textbf{``see less''} but also \textbf{``see inaccurately''}, rendering them unable to sustain fixation on critical visual cues. Collectively, these factors precipitate the loss of fine-grained perception capabilities, thereby motivating our proposed solution.

\section{Related Work}

\subsection{Multimodal Large Language Models}
With the success of large language models (LLMs), extensive efforts have focused on building multimodal large language models (MLLMs) for unified vision-language perception. Most MLLMs adopt a modular architecture, connecting a pretrained visual encoder (e.g., CLIP~\cite{radford2021learning}, SigLIP~\cite{zhai2023sigmoid}) to the LLM via an intermediate projection module. Early models such as BLIP-2~\cite{li2023blip} employ Q-Former based samplers, while later works including Flamingo~\cite{alayrac2022flamingo} integrate visual information through cross-attention within LLM layers. LLaVA~\cite{liu2023visual} and its variants~\cite{zhang2025llava,huang2025hires} popularize a simpler design using lightweight MLPs for visual-text alignment. Despite these advances, existing MLLMs still struggle with fine-grained visual scenarios.

\begin{figure*}[t]
    \centering
    \includegraphics[width=0.98\linewidth]{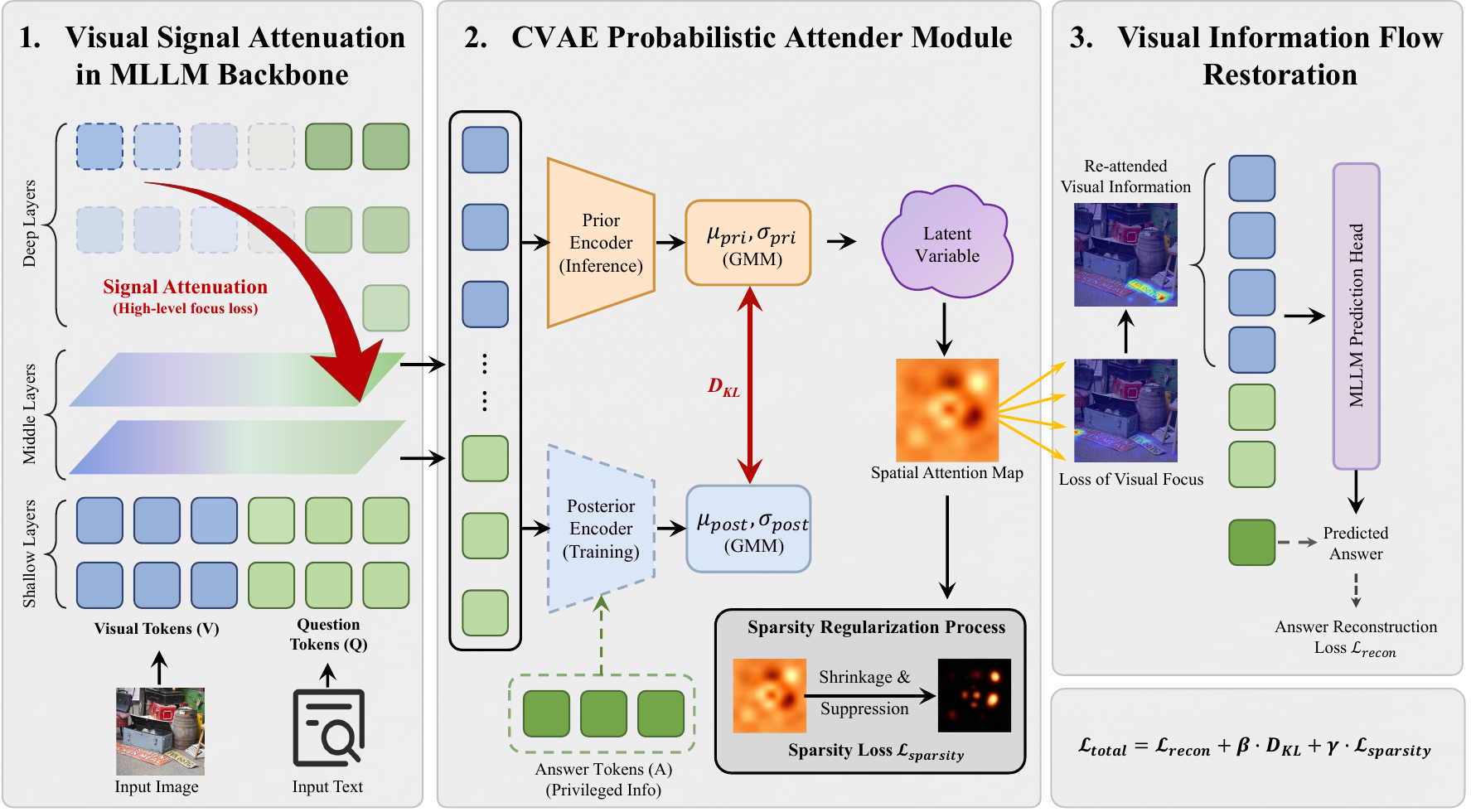}
    \caption{\textbf{Overview of the Variational Information Flow (VIF) Framework.} The framework consists of three stages: \textbf{(1) Visual Signal Attenuation Analysis.} The visual signal is out of focus in the deep layer. The model recovers rich visual cues from intermediate layers. \textbf{(2) CVAE Probabilistic Attender Module.} This module utilizes a GMM-based prior and posterior learning to reconstruct a sparse, task-relevant Spatial Attention Map from middle layers. \textbf{(3) Visual Information Flow Restoration.} This stage recovery involves injecting the learned visual focus into deeper layers to restore fine-grained visual cues to enhance the final prediction.}
    \vspace{-2ex}
    \label{fig:framework}
\end{figure*}

\subsection{Visual Input Optimization in MLLMs}
To alleviate fine-grained visual information loss caused by limited input resolution, many works enhance visual representations via higher-resolution inputs and multi-scale encoding. Early MLLMs process images at fixed low resolutions, constraining fine-grained perception, while later methods increase resolution (e.g., Qwen2.5-VL~\cite{Qwen2.5-VL}, MMFuser~\cite{cao2024mmfuser}) or introduce high-resolution encoders~\cite{guo2024llava}. Another line of work adopts dynamic cropping or patch-based strategies, such as HiRes-LLaVA~\cite{huang2025hires}, and ViCrop~\cite{zhang2025mllms}, to preserve local details through coarse-to-fine re-encoding. While effective at increasing pixel density, these approaches primarily enhance visual inputs and often incur substantial computational overhead, without explicitly addressing how visual information is preserved and exploited during downstream reasoning.

\subsection{Instruction-Aware Visual Guidance}
Beyond input-level improvements, recent works explore guiding visual modeling with textual semantics. Instruction-aware methods argue that different tasks require features from different semantic levels. IGVA~\cite{LI2026111932} re-weights multi-layer visual features conditioned on instructions, while TG-LLaVA~\cite{yan2025tg} employs text-guided modules to suppress irrelevant regions and enhance fine-grained perception.

However, these methods typically rely on deterministic mappings conditioned solely on image and question. For complex fine-grained tasks or weakly specified queries, such guidance can be ambiguous, as multiple regions may satisfy the same instruction. In contrast, our approach models task-relevant visual importance as a probabilistic latent representation and leverages answer supervision during training to resolve ambiguity, enabling more robust visual guidance for multimodal reasoning.
\section{Method}

\subsection{Problem Definition and Overview}

In MLLMs, visual features often suffer severe attenuation through deep layers, as visual tokens (blue) gradually lose influence and are overshadowed by text tokens (green) with higher semantic density (Fig.~\ref{fig:framework}, left). This text-dominated model bias undermines reasoning in the deeper layers, causing the model to lose not only a holistic understanding of the visual scene but also, more critically, its ability to capture fine-grained visual cues.

To mitigate this bottleneck, we propose the \textbf{Variational Information Flow} manipulation framework. The core idea is to model task-relevant visual saliency as a latent probability distribution and to introduce a plug-and-play CVAE-based probabilistic attention module.  This module is designed to assist the backbone model in reconstructing the visual attention distribution within deep layers, re-injecting fine-grained cues into the decision flow to achieve a refocusing on critical visual cues.

\begin{figure}
	\centering
	\includegraphics[width=0.99\linewidth]{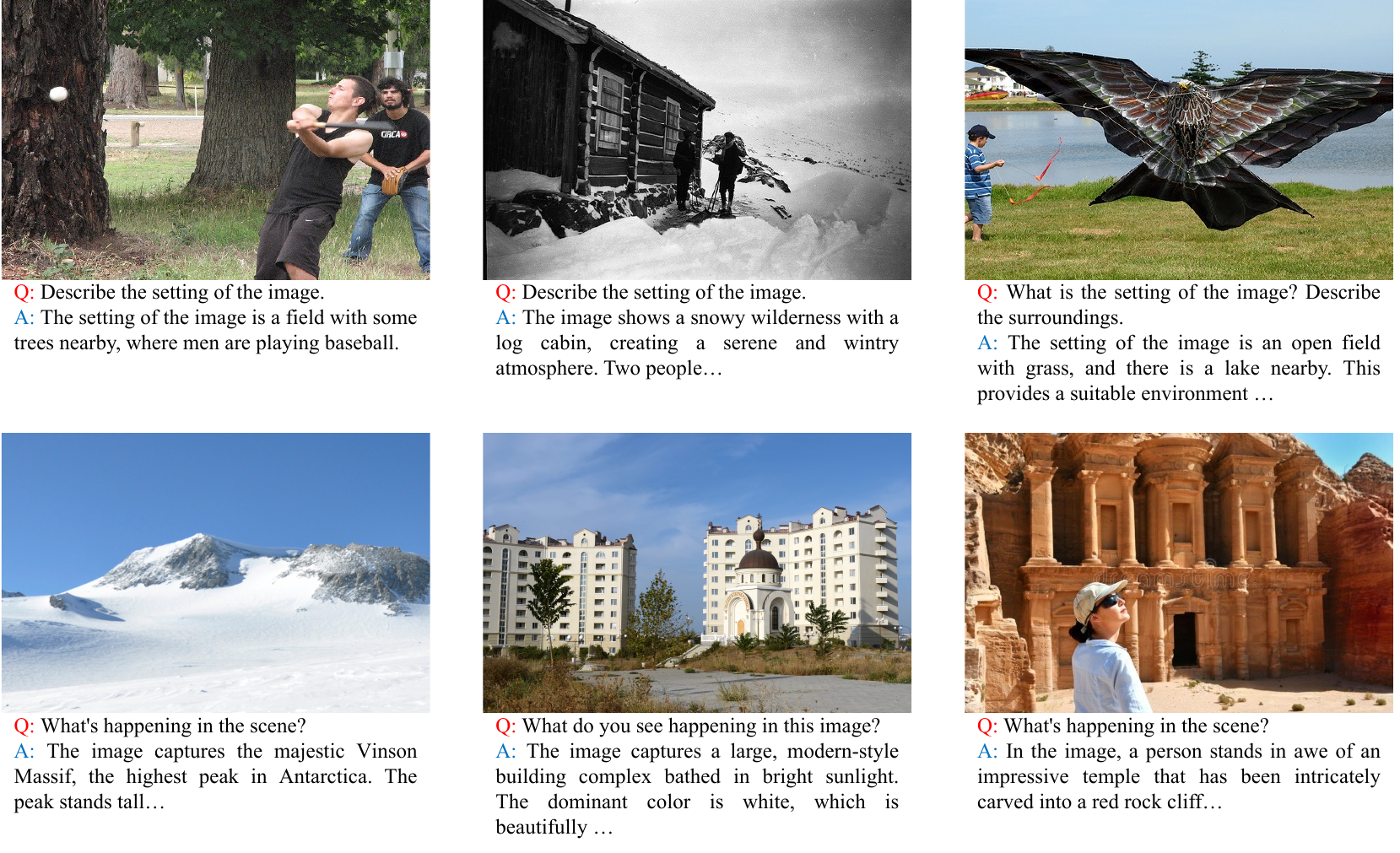}
    \vspace{-2ex}
	\caption{\textbf{Examples of Vision-Language Task Samples.} As illustrated, relying solely on questions often results in semantic ambiguity. Integrating answer information is thus critical to serve as a semantic anchor for robust task-driven visual modeling.}            
    \vspace{-2ex}
	\label{fig:qamotivation}
\end{figure}

\subsection{Probabilistic Modeling via CVAE}

Our goal is to learn a question-conditioned prior $p_\psi(z \mid V, Q)$ that approximates an answer-aware posterior $q_\phi(z \mid V, Q, A)$. After decoding, the latent variable $z$ characterizes the distribution of visual regions that are critical to answering the question under the context of image $V$, question $Q$, and answer $A$.

\textbf{Motivation for Answer-Aware Modeling.}
As shown in Figure~\ref{fig:qamotivation}, relying only on the question $Q$ can lead to semantic ambiguity, because the question alone may not uniquely specify the relevant visual evidence. To mitigate this issue, we introduce the ground-truth answer $A$ during training as an additional semantic anchor that provides reliable supervision for task-relevant visual focus.

We model such uncertain visual importance with a latent variable $z$. Since fine-grained VQA often requires multi-hop reasoning over multiple sparse regions, the sampled latent variables are later decoded into a \textbf{spatial Gaussian mixture model (GMM)} with $K=16$ components, allowing the model to capture multiple candidate visual foci within a unified probabilistic framework.

\textbf{Posterior Learning with Privileged Information (Training Phase).}
During training, the ground-truth answer $A$ is treated as privileged information. We construct the posterior distribution $q_\phi(z \mid V, Q, A)$ to capture the visual cues required to generate the answer $A$. This posterior acts as a teacher that encodes the intrinsic visual dependencies of the task.

\textbf{Prior Inference for Reasoning (Inference Phase).}
During inference, as $A$ is unavailable, we therefore learn a prior distribution $p_\psi(z  \mid V, Q)$ to approximate the posterior. This compels the model to actively infer potential visual foci based solely on the image and question using its inherent reasoning capabilities, thereby achieving a transition from passive perception to active search.

\subsection{Implementation of the Probabilistic Attender Module}
The probabilistic attender consists of three components: latent encoding, spatial GMM decoding, and visual information flow restoration. Its central idea is to transform attenuated visual signals in deep layers into explicit focal cues, thereby turning attenuation into attention. In our formulation, the \emph{information flow} ($\mathcal{IF}$) is explicitly defined as the attention probability distribution of an attention layer, rather than its hidden states.

\textbf{Encoding Latent Variables via CVAE.}
At an middle layer $l$, we extract visual tokens $V$ and question tokens $Q$. During training, answer tokens $A$ are additionally available as privileged information. We build a prior branch and a posterior branch with the same architecture, each consisting of multi-head attention (MHA) and feed-forward networks (FFNs), but with separate parameters.

In each branch, visual and textual tokens first interact via bidirectional cross-attention, and the resulting features are further integrated by an additional MHA layer. The prior branch takes $(V, Q)$ as input, whereas the posterior branch uses the full sequence $[Q, A]$ as textual input to encode answer-aware semantics. The fused representations are then fed into separate linear heads to predict the Gaussian parameters $(\mu_p, \sigma_p^2)$ and $(\mu_q, \sigma_q^2)$, respectively. The latent variables are sampled using the reparameterization trick:
\begin{equation}
	\begin{aligned}
		z_p = \mu_p + \sigma_p \odot \epsilon_p, \\
		z_q = \mu_q + \sigma_q \odot \epsilon_q, \\
		\epsilon_p, \epsilon_q \sim \mathcal{N}(0, \mathbf{I}).
	\end{aligned}
\end{equation}
To encourage the prior distribution available at inference time to approximate the answer-aware posterior, we minimize the KL divergence between the two latent distributions:
\begin{equation}
	\resizebox{0.8\linewidth}{!}{$
	\mathcal{L}_{KL} = \mathbb{E} \left[ D_{KL}\left(\mathcal{N}(\mu_q, \sigma_q^2) \,\|\, \mathcal{N}(\mu_p, \sigma_p^2)\right) \right].
	$}
\end{equation}

\begin{table*}[t]
\centering
\caption{\textbf{Comparison on General Multimodal Capabilities.}  \textbf{Bold} denotes the best result, and \underline{underline} denotes the second best.}
\label{tab:general_results}
\resizebox{\textwidth}{!}{%
\begin{tabular}{l|c|c|ccc|cccc|c|c}
\toprule
\multirow{2}{*}{\textbf{Model}}
& \multirow{2}{*}{\textbf{LLM}}
& \multirow{2}{*}{\textbf{Res.}}
& \multicolumn{3}{c|}{\textbf{Multimodal Understanding}}
& \multicolumn{4}{c|}{\textbf{Knowledge}}
& \textbf{OCR}
& \textbf{Multi-Image} \\
\cmidrule(lr){4-6} \cmidrule(lr){7-10} \cmidrule(lr){11-11} \cmidrule(lr){12-12}
&
&
& \textbf{SEED-I}
& \textbf{LLaVAB}
& \textbf{MME}
& \textbf{GQA}
& \textbf{VQAv2}
& \textbf{AI2D}
& \textbf{SQA}
& \textbf{TextVQA}
& \textbf{BLINK} \\
\midrule
LLaVA-v1.5 & Vicuna-7B & $336\times336$ & 67.2 & 61.8 & 1480.6  & 61.9 & 78.5 & 55.5 & 67.1 & 58.1 & \underline{39.7}  \\
InstructBLIP & Vicuna-7B & $224\times224$& 58.8 & 59.8 & 1137.1  & 49.2 & - & 40.6 & 60.5 & 50.1 & -  \\
mPLUG-Owl2 & LLaMA 2-7B & $448\times448$& 64.5 & 59.9 & 1450.2   & 56.1 & - & 55.7 & 68.7 & 54.3 & -  \\
MiniGPT-v2 & LLaMA 2-7B & $448\times448$& 31.6 & 45.1 & 770.6  & 60.1 & - & 28.4 & 39.6 & - & - \\
LLaMA-Adapter-v2 & LLaMA 2-7B & -& 32.7 & - & 972.7  & - & - & - & - & - & -  \\
IDEFICS-9B & LLaMA 2-7B & -& 45.0 & 45.0 & 942.0  & 38.4 & 50.9 & 42.2 & 53.5 & 25.9 & 38.3 \\
Mantis-8B-Fuyu & Fuyu-8B & $1024\times1024$& 59.3 & 46.8 & 1057.7   & - & - & 46.8 & 56.8 & 49.0 & 38.2 \\
Qwen-VL & Qwen-7B & $448\times448$& 56.3 & 12.9 & 334.1  & 59.3 & {78.8} & 57.7 & 67.1 & {63.8} & 27.9 \\
Qwen-VL-Chat & Qwen-7B & $448\times448$& 65.4 & \textbf{67.7} & 1487.6   & 57.5 & 78.2 & 63.0 & 68.2 & {61.5} & 28.2  \\ 
\midrule

LLaVA-NeXT    & Mistral-7B & $672\times672$ & \underline{69.6} & - & 1519.3 & \textbf{64.2} & - & \textbf{66.6} & 68.5 & \underline{64.9} & -  \\
LLaVA-1.5-HD  & Vicuna-7B  & $672\times1024$ & -   & - & 1414.0 & 54.1 & - & 63.8 & \underline{79.3} & 64.0 & -  \\
Dragonfly    & Vicuna-7B  & $2016\times2016$ & -   & - & 1438.9 & 55.7 & - & \underline{64.2} & \textbf{79.7} & \textbf{66.5} & -  \\
LLaVA-HR     & Vicuna-7B  & $384\times384$& -   & - & \underline{1522.3} & -   & \textbf{80.5} & - & 59.6 & -  & -   \\
\midrule

DenseConnector & Vicuna-7B & - & - & \underline{67.4} & - & \underline{63.8} & - & - & 69.5 & 59.2 & -  \\
MMFuser & Vicuna-7B & - & 60.8 & 65.5 & 1479.7   & 62.8 & - & - & 68.7 & 58.8 & -  \\
LLaVA with ViCrop & Vicuna-7B & $336\times336$ & - & - & - & 60.5 & 75.9 & - & - & 51.7 & -  \\ 

\midrule

IGVA & Vicuna-7B & $336\times336$ & {68.3} & - & {1519.8 }  & 63.1 & - & {57.0} & 70.2 & 59.4 & -  \\
TG-LLaVA & Vicuna-7B & $336\times336$ & 65.0 & - & - & {63.4} & - & - & - & - & - \\
\midrule
\textbf{VIF (Ours)} & Vicuna-7B & $336\times336$ & \textbf{70.8} & 66.4 & \textbf{1547.2 }  & 62.8 & \underline{79.7} & {64.0} & {73.5}  & 59.9 & \textbf{40.5}   \\ 
\bottomrule
\end{tabular}%
}
\end{table*}

\textbf{Decoding to Spatial GMM and Information Flow Injection.}
To capture fine-grained visual focus, we decode the latent set $\{z_k\}_{k=1}^{K}$ into a Spatial GMM. Each component predicts a spatial center $\mu_k^{spa}$, a spread $\sigma_k^{spa}$, and a mixture weight $\pi_k$. On the visual token grid $\{u_n\}_{n=1}^{N}$, the Gaussian response of the $k$-th component is rendered as
\begin{equation}
g_{k,n} = \exp \left( -\frac{\lVert u_n-\mu_k^{spa}\rVert_2^2}{2(\sigma_k^{spa})^2} \right).
\end{equation}
We then aggregate all components into a visual importance map and normalize it into a probability distribution:
\begin{equation}
\resizebox{0.8\linewidth}{!}{$
	V_{Map_n} = \sum_{k=1}^{K} \pi_k g_{k,n}, 
	\hat{V} = \mathrm{Softmax}(V_{Map}).
$}
\end{equation}

Motivated by the analysis in introduction, where middle layers preserve stronger localization ability while deep layers often suffer from visual defocusing, we adopt a \textit{selective layer patching} strategy. Specifically, we perform pair-wise injection from middle layers $l \in \{11, 13, 15, 17\}$ to deep layers $l' \in \{25, 27, 29, 31\}$. For each injection layer $l'$, the original information flow is defined as the standard attention probability distribution:
\begin{equation}
\resizebox{0.8\linewidth}{!}{$
\mathcal{IF}_{ori}^{(l')} = \mathrm{Softmax}\left(\frac{Q^{(l')}(K^{(l')})^\top}{\sqrt{d_h}} + M\right),
$}
\end{equation}
where $M$ is the visibility mask. We then inject the decoded visual importance as an explicit focal bias:
\begin{equation}
\widetilde{\mathcal{IF}}_{inj}^{(l')} = \mathcal{IF}_{ori}^{(l')} + \alpha \cdot \hat{V},
\end{equation}
where $\alpha$ controls the injection strength. 

To preserve the attention constraints on invisible regions, we apply masked re-normalization to obtain the final injected information flow:
\begin{equation}
\mathcal{IF}_{inj}^{(l')} = \mathrm{Norm}\Big(\widetilde{\mathcal{IF}}_{inj}^{(l')} \odot \mathbf{1}_{visible}\Big),
\end{equation}
where $\mathrm{Norm}$ denotes row-wise normalization such that each attention distribution sums to 1. In the selected deep layer, $\mathcal{IF}_{inj}^{(l')}$ is used in place of the original attention probability matrix to compute the subsequent attention output, while all other operations remain unchanged.

\subsection{Joint Optimization Objective}
Our training objective follows the standard variational inference paradigm, aiming to maximize the Evidence Lower Bound (ELBO) of the conditional marginal likelihood $\log p_\theta(A|V, Q)$.

\textbf{Derivation of the Variational Lower Bound.}
By introducing a variational posterior $q_\phi(z|V,Q,A)$ to approximate the true posterior, and applying Jensen's inequality, the ELBO is derived as follows:
\begin{equation}
\resizebox{0.95\linewidth}{!}{$
\begin{aligned}
\log p_\theta(A \mid V, Q) 
& \ge \mathbb{E}_{z \sim q_\phi} \left[ \log \frac{p_\theta(A \mid V, Q, z) \, p_\psi(z \mid V, Q)}{q_\phi(z \mid V, Q, A)} \right] \\
& = \mathbb{E}_{z \sim q_\phi} [\log p_\theta(A \mid V, Q, z)] \\
& \quad - D_{\text{KL}}(q_\phi(z \mid V, Q, A) \parallel p_\psi(z \mid V, Q))
\end{aligned}
$}
\end{equation}
where the first term represents the response reconstruction objective aiming to maximize the likelihood of the ground-truth answer, optimized by the MLLM backbone parameters $\theta$, while the second term enforces consistency between the variational posterior (parameterized by $\phi$) and the conditional prior (parameterized by $\psi$).

\textbf{Sparsity Regularization.}
To prevent the model from converging to trivial solutions that uniformly cover the entire image, we impose sparsity constraints on the Gaussian renderer. Specifically, we constrain the ``volume'' of each Gaussian component, defined as the product of its amplitude and variance, thereby encouraging the model to emphasize only the most critical regions. In addition, we incorporate an entropy regularizer to form a comprehensive sparsity objective, defined as:
\begin{equation}
\mathcal{L}_{\text{sparsity}} = \mathcal{H}(\boldsymbol{\pi})
+ \frac{1}{K} \sum_{k=1}^{K} (\pi_k \times \boldsymbol{\sigma}_k^2),
\end{equation}
where $\mathcal{H}(\boldsymbol{\pi}) = - \sum_{k=1}^{K} \pi_k \log \pi_k$ denotes the Shannon entropy, promoting sparse mixture distribution. Combined with this entropy regularization, the volume term simultaneously suppresses the variance $\boldsymbol{\sigma}_k$ (shrinking the spatial extent) and the amplitude $\pi_k$ (dampening uncertain activations).

\textbf{Total Loss Function.}
The final joint optimization objective is defined as:
\begin{equation}
\mathcal{L}_{\text{total}} = \mathcal{L}_{\text{recon}} + \beta \mathcal{L}_{KL} + \gamma \mathcal{L}_{\text{sparse}}
\end{equation}
where $\mathcal{L}_{\text{recon}} = -\log p_\theta(A|V,Q,z)$ denotes the standard cross-entropy loss for answer generation, and $\beta$ and $\gamma$ are hyper-parameters used to balance the three terms.

\section{Experiments}

\begin{table}[t]
\centering
\caption{\textbf{Fine-Grained Perception and Referring Expression Comprehension.} \textbf{Bold} denotes the best result, and \underline{underline} denotes the second best. REC results are reported with the CIDEr score.}
\label{tab:finegrained_results}
\resizebox{\columnwidth}{!}{%
\begin{tabular}{l|cc|c|ccc}
\toprule
\multicolumn{1}{c|}{\multirow{2}{*}{\textbf{Model}}} &
\multicolumn{2}{c|}{\textbf{HR-Bench}} &
\multirow{2}{*}{\textbf{Vstar}} &
\multicolumn{3}{c}{\textbf{REC (RefCOCO)}} \\
\cmidrule(lr){2-3} \cmidrule(lr){5-7}
 & \textbf{4K} & \textbf{8K} &  &
 \textbf{val} & \textbf{testA} & \textbf{testB}  \\
 \midrule
LLaVA-v1.5 & 36.1 & 32.1 & 45.0 & 30.4 & 16.0 & 42.0 \\
mPLUG-Owl2 & 36.9 & \underline{33.8} & 35.6 & - & - & - \\
MiniGPT-v2 & 25.5 & 26.1 & - & - & - & -  \\
IDEFICS-9B & 30.6 & 28.8 & - & - & - & -  \\
Qwen-VL & 31.8 & 28.4 & - & - & - & - \\
IGVA & \underline{40.0} & - & \underline{48.2} & - & - & - \\
MMFuser & - & - & - & \underline{33.6} & \textbf{17.7} & \underline{45.9} \\
LLaVA with ViCrop & - & - & 46.1 & - & - & - \\ \midrule
\textbf{VIF (Ours)} & \textbf{44.8} 	& \textbf{36.8} 	& \textbf{50.8} 	& \textbf{34.3} 	& \underline{16.6} & \textbf{47.0} \\ 
 \midrule
 \midrule
Qwen2.5-VL-7B & 68.6  & 64.9 & 76.4 & - & -& - \\
CoVT & \underline{71.0} & \underline{68.6} & \underline{77.5} &- &- & - \\
\textbf{VIF w CoVT (Ours)} & \textbf{71.5} & \textbf{68.8} & \textbf{79.0} & -&- & - \\
\bottomrule
\end{tabular}%
}
\end{table}

To systematically validate the effectiveness of our proposed VIF framework in restoring fine-grained visual information, enhancing grounding capabilities, and maintaining general multimodal understanding, we conducted studies across 12 diverse benchmarks, spanning 15 distinct evaluation tasks.
All experiments are conducted at the 7B scale, a controlled and widely adopted setting for MLLM. Under this setup, we train and validate VIF on a 7B-class backbone and benchmark it extensively across a broad range of multimodal tasks, with direct comparisons against multiple strong 7B-scale baselines. The results consistently demonstrate the effectiveness and robustness of our approach.

\begin{table}[t]
\centering
\caption{\textbf{Ablation Study of the VIF Framework.}}
\label{tab:ablation_vif}
\resizebox{\columnwidth}{!}{%
\begin{tabular}{l|c|c|cc|c|ccc}
\toprule
\multicolumn{1}{c|}{\multirow{2}{*}{\textbf{Model}}} &
\multirow{2}{*}{\textbf{SEED-I}} &
\multirow{2}{*}{\textbf{AI2D}} &
\multicolumn{2}{c|}{\textbf{HR-Bench}} &
\multirow{2}{*}{\textbf{Vstar}} &
\multicolumn{3}{c}{\textbf{REC (RefCOCO)}} \\
\cmidrule(lr){4-5} \cmidrule(lr){7-9}
& & & \textbf{4K} & \textbf{8K} &  &
 \textbf{val} & \textbf{testA} & \textbf{testB}  \\
 \midrule
VIF (Ours)       & \textbf{70.8} & \textbf{64.0} & \textbf{44.8} & \textbf{36.8} & \textbf{50.8} & \textbf{34.3} & \textbf{16.6} & \textbf{47.0} \\
- w/o AP           & 69.2 & 63.0 & 42.3 & 34.3 & 48.7 & 32.8 & 15.3 & 44.1 \\
- w/o SP           & 68.2 & 62.5 & 36.8 & 32.9 & 49.2 & 29.6 & 13.9 & 40.7 \\
Full-Seq         & 68.5 & 62.8 & 38.9 & 33.6 & 49.2 & 29.9 & 14.1 & 42.7 \\
Deep-Only        & 68.2 & 63.1 & 39.1 & 33.0 & 49.7 & 30.7 & 15.6 & 43.1 \\
Mid-Deep Feature & 68.3 & 63.2 & 39.3 & 32.5 & \textbf{50.8} & 29.8 & 14.3 & 41.7 \\
\hline
\end{tabular}
}
\end{table}

\subsection{Datasets and Evaluation Metrics}
To comprehensively evaluate model performance, we organize benchmarks into 4 dimensions.

\textbf{(1) Multimodal Understanding.}
We evaluate general perception and cognition using MME~\cite{fu2025mme} and SEED-Bench (Image)~\cite{li2024seed}, instruction-following with LLaVA-Bench~\cite{liu2023visual}, and multi-image robustness with BLINK~\cite{fu2024blink}.
\textbf{(2) Knowledge-Intensive and Logical Reasoning.}
We adopt ScienceQA~\cite{lu2022learn} and VQAv2~\cite{jia2024vqa} as core VQA benchmarks, complemented by GQA~\cite{ainslie2023gqa} for compositional spatial reasoning and AI2D~\cite{kembhavi2016diagram} for scientific diagram understanding.
\textbf{(3) OCR.}
TextVQA~\cite{singh2019towards} is used to assess optical character recognition on text-rich images.
\textbf{(4) Fine-Grained Perception and Referring Expression Comprehension.}
We use HR-Bench~\cite{hrbench} and V\textsc{star}~\cite{wu2023v} to evaluate fine-grained visual perception, and RefCOCO~\cite{yu2016modeling} to assess visual grounding and referring expression comprehension.

\subsection{Baselines}
To evaluate our method, we compare it with representative MLLMs, grouped as follows:
\textbf{General Baselines:} LLaVA-v1.5~\cite{liu2023visual}, InstructBLIP~\cite{dai2023instructblip}, mPLUG-Owl2~\cite{ye2023mplug}, Qwen-VL~\cite{Qwen-VL}, Qwen-VL-Chat~\cite{Qwen-VL}, LLaMA-Adapter-v2~\cite{gao2023llama}, IDEFICS-9B~\cite{laurencon2023obelics}, Mantis-8B-Fuyu~\cite{Jiang2024MANTISIM}, and MiniGPT-v2~\cite{chen2023minigpt}.
\textbf{High-Resolution Models:} LLaVA-NeXT~\cite{liu2024llavanext}, LLaVA-1.5-HD~\cite{liu2023improvedllava}, Dragonfly~\cite{thapa2024dragonfly}, and LLaVA-HR~\cite{luo2024feast}.
\textbf{Feature-Enhanced Models:} DenseConnector~\cite{yao2024dense}, MMFuser~\cite{cao2024mmfuser}, LLaVA with ViCrop~\cite{zhang2025mllms}, IGVA~\cite{LI2026111932}, and TG-LLaVA~\cite{yan2025tg}.
We also include Qwen2.5-VL-7B~\cite{Qwen2.5-VL} and its Chain-of-Thought (CoT) fine-tuned variant, CoVT~\cite{qin2025chain}, and build our method on top of the CoVT backbone. For fair comparison, we strictly follow CoVT's original training protocol, using Qwen2.5-VL-7B with LoRA fine-tuning (rank 16, alpha 32), four training phases (4000/3000/3000/5000 steps), and the same vision-centric real-world and spatial perception data; because our fine-tuning dataset (LLaVA-LAION/CC/SBU) is smaller and lower quality than the closed-source data used by Qwen2.5-VL-7B, we adopt CoVT's CoT fine-tuning recipe instead of directly fine-tuning on our dataset.

\begin{figure*}[t]
  \centering
  \includegraphics[width=0.98\linewidth]{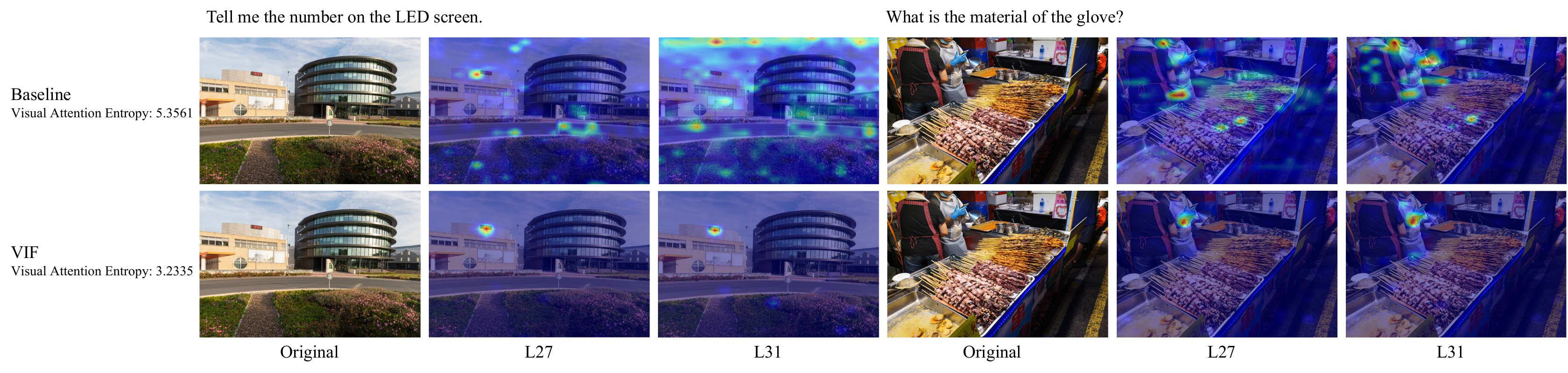}
  \vspace{-2ex}
  \caption{\textbf{visualization of attention maps comparing the baseline and our proposed model.} The top and bottom rows display the attention distributions of the baseline and our model, across different layers (L27 and L31).}
  \label{fig:visualization}
\end{figure*}

\begin{table*}[t]
\centering
\caption{\textbf{Ablation study on {extraction to injection} layer configurations.}}
\label{tab:layer_mapping_ablation}
\resizebox{0.95\textwidth}{!}{
\begin{tabular}{l|l|l|c|c|c|c|c}
\toprule
\textbf{Method / Configuration} & \textbf{Extraction Layers} & \textbf{Injection Layers} & \textbf{SEED-I} & \textbf{GQA} & \textbf{TextVQA} & \textbf{HR-Bench (4K)} & \textbf{Vstar} \\
\midrule
LLaVA-v1.5 (Baseline)      & --                & --                & 67.2 & 61.9 & 58.1 & 36.1 & 45.0 \\
VIF (L-early, 4 pairs)     & 9,11,13,15        & 25,27,29,31       & 69.5 & 62.3 & 59.1 & 42.5 & 48.9 \\
VIF (L-late, 4 pairs)      & 13,15,17,19       & 25,27,29,31       & 70.3 & 62.6 & 59.5 & 43.8 & 49.6 \\
VIF (5 pairs)              & 9,11,13,15,17     & 23,25,27,29,31    & \textbf{70.9} & \textbf{62.9} & 59.8 & \textbf{45.0} & \textbf{50.9} \\
\midrule
\textbf{VIF (Our Default)} & \textbf{11,13,15,17} & \textbf{25,27,29,31} & 70.8 & 62.8 & \textbf{59.9} & 44.8 & 50.8 \\
\bottomrule
\end{tabular}
}
\end{table*}

\subsection{Experimental Results}

\subsubsection{General Multimodal Capability}

As shown in Table~\ref{tab:general_results}, VIF achieves competitive performance across a broad range of benchmarks. Compared with models of similar scale and settings, VIF shows consistent gains on most tasks, indicating its effectiveness and robustness.

\textbf{Comparison with General Baselines.}
VIF performs favorably against standard 7B-scale models. Relative to its backbone LLaVA-v1.5, VIF achieves clear improvements on multiple benchmarks, including ScienceQA and LLaVA-Bench, and attains the best result on the multi-image benchmark BLINK, suggesting improved handling of fine-grained visual evidence. Compared with InstructBLIP and Qwen-VL-Chat, VIF exhibits more stable performance across multimodal tasks.

\textbf{Comparison with High-Resolution Models.}
Despite using a standard input resolution of $336 \times 336$, VIF achieves results comparable to high-resolution approaches with substantially larger visual sequences. While ultra-high-resolution models retain advantages on pixel-level OCR, VIF demonstrates strong overall performance on comprehensive benchmarks such as MME and SEED-I, reflecting a favorable efficiency–performance trade-off.

\textbf{Comparison with Feature Enhancement Models.}
VIF consistently outperforms existing feature enhancement methods such as IGVA and TG-LLaVA on comprehensive benchmarks. On SEED-I, VIF surpasses both baselines, suggesting that variational information flow is more effective than prior feature fusion or connector-based strategies for preserving task-relevant visual information.

\subsubsection{Fine-Grained Perception and Referring Expression Comprehension}
Table~\ref{tab:finegrained_results} summarizes results on fine-grained perception and referring expression comprehension. On \textbf{HR-Bench} (4K/8K) and \textbf{V\textsc{star}}, VIF consistently outperforms the backbone LLaVA-v1.5 and feature-enhanced methods such as IGVA. For example, on V\textsc{star}, VIF achieves 50.8, surpassing IGVA (48.2), demonstrating improved sensitivity to tiny objects and subtle visual cues. For visual grounding, VIF also outperforms MMFuser on RefCOCO (val/testB), indicating enhanced spatial reasoning and referring expression comprehension.

\textbf{Generalization to Stronger Backbones.}
To examine the scalability of our approach, we further apply VIF to the advanced Qwen2.5-VL-7B backbone. Under the CoVT setting, although CoVT already exhibits strong reasoning performance, \textbf{VIF w/ CoVT} still yields consistent gains on fine-grained benchmarks. This suggests that VIF remains effective when paired with stronger backbones and advanced reasoning strategies. Moreover, this result helps position VIF with respect to recent ``thinking with images'' and visual chain-of-thought paradigms. While these approaches typically improve performance by enriching the input signal through iterative visual querying or external tools, VIF focuses on a complementary direction by strengthening how the MLLM backbone utilizes the already-provided visual inputs. The gains achieved by \textbf{VIF w/ CoVT} therefore indicate that VIF can serve as a complementary enhancement to stronger visual reasoning paradigms.

\subsubsection{Ablation Study}

Table~\ref{tab:ablation_vif} summarizes our ablation studies on the VIF framework. We evaluate the impact of: (1) removing answer posterior supervision; (2) omitting sparsity regularization; (3) applying importance scores across the full token sequence; (4) focusing on deep layers only ($l = l' \in {25, 27, 29, 31}$); and (5) injecting middle-layer visual features into deep-layer representations.

\textbf{Impact of Supervision and Regularization.}
Removing answer posterior supervision (\textit{w/o AP}) consistently degrades performance (e.g., a 1.6\% drop on SEED-I), confirming its role in guiding the model to focus on task-relevant visual regions. Omitting sparsity regularization (\textit{w/o SP}) sharply reduces fine-grained performance: HR-Bench (4K) drops from 44.8 to 36.8, and REC (val) falls from 34.3 to 29.6. These results highlight the importance of sparsity in filtering noise and maintaining compact, discriminative information flow.

\textbf{Scope and Layer Selection Analysis.}
Applying importance weighting to the full token sequence (\textit{Full-Seq}) worsens results due to noise from indiscriminate enhancement. The \textit{Deep-Only} variant underperforms, as deep-layer textual tokens already capture coarse visual summaries, making them less suitable for learning informative visual importance. The \textit{Mid-Feature} strategy is less effective than our design, emphasizing the need to model attention flow rather than injecting visual features.

\textbf{Impact of Extraction to Injection Layer Mapping.}
We compare different extraction to injection layer configurations, including earlier vs.\ later extraction layers and 4-pair vs.\ 5-pair mappings. Extracting from relatively early layers (\textit{L-early, 4 pairs}) leads to slightly worse results across benchmarks. By contrast, using middle extraction layers and injecting into deeper layers performs more favorably, which validates our design choice of restoring visual information from intermediate representations into high-level reasoning layers. Although the 5-pair variant achieves slightly better results on several benchmarks, its gains over our default 4-pair setting are marginal. We therefore adopt the 4-pair mapping $[11,13,15,17]\rightarrow[25,27,29,31]$ as the default, as it provides a better balance between performance and computational cost.

\subsubsection{Deep Layers Attention Analysis}
Our approach was validated through both qualitative and quantitative analysis.  As illustrated in Figure \ref{fig:visualization}, unlike the baseline, which suffers from attention dispersion due to background noise (e.g., sky) or irrelevant objects (e.g., food), our model exhibits superior focus in deep layers (L27, L31). It effectively locates fine-grained targets, such as the small LED screen or the glove. Statistical analysis on 500 randomly sampled instances shows that our model reduces visual attention entropy from 5.3561 (baseline) to 3.2335, indicating improvements in both attention sparsity and certainty. These results further suggest that VIF helps the model form more concentrated and reliable task-relevant visual grounding in the deep layers.
\section{Conclusion}

In this work, we identified and addressed the critical bottleneck of \textbf{Visual Attenuation} in MLLMs, where fine-grained visual cues are progressively discarded and attention structures collapse in the deep layers. To address this challenge, we propose the \textbf{Variational Information Flow (VIF)} framework, a novel paradigm that shifts from passively augmenting input features to actively reconstructing deep-layer information flow. By leveraging a CVAE-based posterior learning mechanism together with a Gaussian Mixture Model (GMM) prior, VIF effectively reconstructs task-relevant visual signals while enforcing spatial sparsity.

\section*{Limitations}

While our method effectively enhances fine-grained visual perception, it has two main limitations. First, although the proposed CVAE module is designed to be lightweight, it inevitably introduces additional parameters and computational overhead compared to the original backbone during inference. Future work that enables the model itself to maintain focused attention on critical visual cues could improve efficiency while enhancing performance. Second, since our approach operates at the level of information flow to restore attenuated features, its performance is partially constrained by the original input resolution of the frozen visual encoder (e.g., CLIP). For extremely small objects whose pixel-level information is lost during initial encoding, feature reconstruction remains challenging.

\section*{Acknowledgments}
We thank the anonymous reviewers for their constructive comments and suggestions, which helped improve this manuscript. This work was supported by the Beijing Natural Science Foundation under Grant No. L257006.

\bibliography{custom}


\end{document}